\documentclass[twocolumn,10pt]{asme2ef}

\usepackage[utf8]{inputenc}
\usepackage[T1]{fontenc}
\usepackage{graphicx}
\usepackage{grffile}
\usepackage{longtable}
\usepackage{wrapfig}
\usepackage{rotating}
\usepackage[normalem]{ulem}
\usepackage{amsmath}
\usepackage{textcomp}
\usepackage{amssymb}
\usepackage{capt-of}
\usepackage{hyperref}
\usepackage{graphicx,psfrag,amsmath,amssymb}
\usepackage{url,color, xcolor}
\usepackage[ruled, lined, linesnumbered]{algorithm2e}
\usepackage[hyperpageref]{backref}
\hypersetup{
colorlinks   = true, 
urlcolor     = magenta, 
linkcolor    = red, 
citecolor   = {green!50!black} 
}
\usepackage[caption=false]{subfig}

\newcommand{\bp}{\boldsymbol{p}}

\newcommand{\bn}{\boldsymbol{n}}
\newcommand{\bc}{\boldsymbol{c}}

\usepackage{xspace}
\makeatletter
\DeclareRobustCommand\onedot{\futurelet\@let@token\@onedot}
\def\@onedot{\ifx\@let@token.\else.\null\fi\xspace}
\def\eg{\emph{e.g}\onedot}

\def\etal{\emph{et al}\onedot}
\author{Guoxiang Zhang
\affiliation{
Mechatronics, Embedded Systems and Automation Lab\\
School of Engineering\\
University of California, Merced\\
Merced, California 95348\\
Email: gzhang8@ucmerced.edu
}
}
\author{YangQuan Chen\thanks{Address all correspondence to this author.}
\affiliation{
Mechatronics, Embedded Systems and Automation Lab\\
School of Engineering\\
University of California, Merced\\
Merced, California 95348\\
Email: ychen53@ucmerced.edu
}
}
\title{Self-optimizing loop sifting and majorization for 3D reconstruction}
\begin{document}

\maketitle
\begin{abstract}
\emph{Visual simultaneous localization and mapping (vSLAM) and 3D reconstruction methods have gone through impressive progress. These methods are very promising for autonomous vehicle and consumer robot applications because they can map large-scale environments such as cities and indoor environments without the need for much human effort. However, when it comes to loop detection and optimization, there is still room for improvement. vSLAM systems tend to add the loops very conservatively to reduce the severe influence of the false loops. These conservative checks usually lead to correct loops rejected, thus decrease performance. In this paper, an algorithm that can sift and majorize loop detections is proposed. Our proposed algorithm can compare the usefulness and effectiveness of different loops with the dense map posterior (DMP) metric. The algorithm tests and decides the acceptance of each loop without a single user-defined threshold. Thus it is adaptive to different data conditions. The proposed method is general and agnostic to sensor type (as long as depth or LiDAR reading presents), loop detection, and optimization methods. Neither does it require a specific type of SLAM system. Thus it has great potential to be applied to various application scenarios. Experiments are conducted on public datasets. Results show that the proposed method outperforms state-of-the-art methods.}
\end{abstract}

\section*{INTRODUCTION}
\label{sec:org6b55b35}

Due to rapid development in autonomous vehicles and consumer robots, there is an increasing need for precise 3D maps for route and action planning and navigation. Among 3D mapping methods, visual simultaneous localization and mapping (vSLAM) and 3D reconstruction methods are very promising because they can map large-scale environments such as cities and indoor environments without the need for much human effort involved. 

vSLAM and 3D reconstruction methods have gone through impressive progress. In camera tracking, there are different methods, such as sparse keypoint point-based methods~\cite{MurArtal2017,Endres2014,Klein2007,Dai2017e,Liang2019}, direct methods \cite{Zubizarreta2020,Engel2014}, and dense surface-based methods~\cite{Newcombe2011,Whelan2015}. Additionally, IMU are added to methods~\cite{Concha2016,Usenko2016,Niesner2014,ZihaoZhu2017,Dong2017,Liu2018b,Concha2016} to make tracking more accurate. Even though camera tracking algorithms have good performance and low drift, the build-up error can still not be ignored~\cite{Cadena2016}. To solve this problem, loop closure detection~\cite{GalvezLopez2012,Cummins2008} and optimization~\cite{Kummerle2011} are often leveraged to counter the problem, and it has provided plenty of improvements. However, the problem is not fully solved yet. Intuitively, the more loops in data, the more information to recover more precise camera trajectories and 3D models. But, in practice, when running existing vSLAM systems on datasets with loopy motions, mismatches can always be found in the final results. This means that loops are not successfully detected and utilized.

vSLAM systems tend to add the loops very conservatively to reduce the severe influence of the false loops~\cite{Cadena2016}. These conservative checks are the result of the non-perfect precision performance of loop detection methods. There are high chances that detected loops are incorrect ones.

To solve this challenging problem, we propose an algorithm that can sift and majorize loop detections so that only correct and essential loops are fed into the following optimization steps. The proposed method highly couples with the dense map posterior (DMP) metric~\cite{Zhang2021a} that can evaluate 3D reconstruction performance without ground truth measurement. Our proposed algorithm can compare the usefulness and effectiveness of different loops and ultimately sifts out false and unimportant loops. To the best of our knowledge, the contributions of the proposed algorithm are:
\begin{enumerate}
\item The proposed algorithm can sift loop detections based on their impact on loop optimization results.
\item It is the first algorithm that can marjorize loop detection only to keep the important ones while ignoring the less relevant ones.
\item Experiments on public datasets show it outperforms state-of-the-art methods.
\end{enumerate}

\section*{RELATED WORK}
\label{sec:org46ab58a}

To avoid the severe consequence of optimizing with false loops, vSLAM and 3D mapping systems tend to add the loops very conservatively. ORB-SLAM2~\cite{MurArtal2017} requires the presence of several consistent loops in consecutive keyframes to accept them, where at least one keyframe must be shared in order to be classified as consistent. With this consistency check, ORB-SLAM2 merely takes false loops into optimization but at the price that plenty of correct loops are rejected. ElasticFusion~\cite{Whelan2015} evaluates several characteristics before taking a loop detection into optimization pipelines, including deformation cost and final state of the Gauss-Newton system. Even after all the evaluations, a good loop is often rejected, and not rare to see that an incorrect loop is accepted. BundleFusion~\cite{Dai2017e} filters loop correspondences with cascade checks including local geometric and photometric consistency checks and check on correspondence residual after optimization. The local depth discrepancy check shares a small similarity with our work. However, the check is limited to a very local region with a downsampled depth resolution together with a user-specified threshold. Thus it is not informed about the effect of loop data correspondences impacting a full 3D model. A requirement on a user parameter also makes it ineffective and less adaptive. \cite{Endres2014} do this by pruning edges after optimization based on the discrepancy between the individual transformation estimates before and after optimization. We share the idea of observing optimization consequences brought in with a loop, but their impact is measured on a sparse graph while ours is observed on a full 3D dense model.

Another approach to solving the problem is to treat false loops as outlier data and decrease their impact on the optimization~\cite{Lee2013,Sunderhauf2013,Agarwal2013,Sunderhauf2012}. They work well in some cases, but the dependence on initial conditions and the ratio of outliers makes them prune to failures.  
Choi \etal further develop this idea into an algorithm that is highly coupled with the dense 3D reconstruction problem by specifying both pose graph construction and least square information calculation~\cite{Choi2015}. This method is very effective when a desired camera scan pattern is followed but it requires keeping surface within camera range all the time thus limiting its flexibility. It also suffers dependence on initial condition and outlier ratio.

Due to the difficulty of balancing precision and recall of loop detections, SUN3D~\cite{Xiao2013} turns to a human-in-the-loop approach by labeling objects in scenes and connect the same objects across frames. This method performs very well in terms of loop precision and recall, but it requires too much effort in labeling; thus is not practical to process data on a large scale. 

\section*{METHOD}
\label{sec:org42b5eaa}
To solve the loop sifting problem, we propose an algorithm specified in Algorithm~\ref{algo1}. In the algorithm, a given set of loop detections is denoted as \(\mathbb{O}\) among which each individual one is denoted as \(O\). The supporting optimization pose graph is denoted as \(\mathcal{G}\). The sensor (\eg camera and LiDAR) data are denoted as \(\mathbb{Z}\). 

There are two parts in the algorithm. In the first part, all the loops are tested and evaluated individually on the given initial pose graph. This step first runs optimization with a single loop and then fuses a model with the optimized results \(\mathbb{T}^i\). Then a DMP value \(r\) is evaluated for the fused 3D model \(M\). This means that it tests each loop and sees how much improvement it provides by itself. Finally, all these loops are ranked by the calculated DMP value \(r\) in ascending order (more effective \(\rightarrow\) less effective \(\rightarrow\) negative impacts). 

In the second part, all the loops are tested and evaluated one more time, but in a way that is different from the first time. In this part, the loops are tested in sorted order: the ones that provide more improvements are tested first. When a loop can provide performance improvement on the previous result, it will be added to an accepted set, thus will also impact consequent loop tests. In this way, loops are accepted when they can provide performance improvement on the current status. The first accepted one should make an improvement to the original results from tracking.

\begin{algorithm}
    \SetKwFunction{findInliers}{findInliers}
    \SetKwFunction{rank}{rank}
    \SetKwFunction{r}{r}
    \SetKwFunction{opt}{optimize}
    \SetKwFunction{len}{len}
    \SetKwFunction{union}{union}
    \SetKwFunction{fuseModel}{fuseModel}

    \SetKwInOut{KwIn}{Input}
    \SetKwInOut{KwOut}{Output}

    \KwIn{ $\mathbb{O}$, $\mathcal{G}$, $\mathbb{Z}$ } 
    \KwOut{filtered loops $\mathbb{O}^*$}

    $\mathbb{O}^* \leftarrow \varnothing$, $r^* \leftarrow \r(\mathcal{G}, \mathbb{Z})$, $\boldsymbol{r} \gets \varnothing$

    \For{$i \gets 0$ \KwTo $\len(\mathbb{O})$}{
        $\mathbb{T}^i \gets \opt(\mathcal{G}, \mathbb{O}[i])$
        
        $M^i \gets \fuseModel(\mathbb{T}^i, \mathbb{Z})$

        $\boldsymbol{r}[i] \gets \r(M^i, \mathbb{Z})$

    }
    \tcp{Ranking $\mathbb{O}$ by our metric}    
    $\mathbb{O}^r \leftarrow \rank(\mathbb{O},\: by=\boldsymbol{r},\: order=\text{descending})$ 
    
    \tcp{Try $\mathbb{O}^r$ one by one and add the ones making improvements}

    \For{$i \gets 0$ \KwTo $\len(\mathbb{O}^r)$}{
        $\mathbb{O}^*_{tmp} \gets \union(\mathbb{O}^*, \mathbb{O}[i])$ 

        $\mathbb{T}^\prime \gets \opt(\mathcal{G}, \mathbb{O}^*_{tmp})$
        
        $M^\prime \gets \fuseModel(\mathbb{T}^\prime, \mathbb{Z})$

        $r^\prime \gets \r(M^\prime, \mathbb{Z})$

        \If{$r^\prime > r^*$}{
            $\mathbb{O}^* \gets \mathbb{O}^*_{tmp}$

            $r^* \gets r^\prime$
         }

    }
    \caption{Loop sifting and majorization algorithm}
    \label{algo1}
\end{algorithm}

\section*{IMPLEMENTATION}
\label{sec:org531bedd}
The proposed method is general and agnostic to loop detection and optimization methods. Neither does it require a specific type of vSLAM system. For our experiments, we choose several well know implementations.

\subsection*{Tracking and optimization pose graph}
\label{sec:org46f2ce0}

The proposed method requires an optimization pose graph as input data. The only requirement of the pose graph optimization is that it can handle loop closure optimization. In our implementation, we use sparse image feature-based tracking and mapping method implemented by ORB-SLAM2~\cite{MurArtal2017} with loop detection disabled. The pose graph from ORB-SLAM2 is utilized as the optimization graph for the proposed method. For the purpose of loop sifting and majorization, we find pose graph optimization is good enough; thus, the more time-consuming full bundle adjustment is not included.  

The ORB-SLAM2 is a very well implemented sparse feature-based SLAM system.  Inside this tracking module, the Oriented FAST and Rotated BRIEF (ORB) features are extracted for keypoint matching. Then frames are tracked against keyframes with motion estimate and then refined with a local sparse map. Keyframes are generated when tracking is weak, or the local bundle adjustment thread is free. Local BA is used to correct the re-projection error of feature correspondences among co-visible keyframes in a background thread. This tracking module provides camera poses for each frame and a co-visibility graph across keyframes.

\subsection*{Model fusion}
\label{sec:org1a7c444}
It is a important step to fuse camera reading data into dense 3D models . For this step, surfels~\cite{Pfister2000} are used as a data representation of 3D model. Each surfel has seven attributes: a position \(\bp \in \mathbb{R}^3\), normal \(\bn \in \mathbb{R}^3\), color \(\bc \in \mathbb{N}^3\), weight \(w \in \mathbb{R}\), radius \(r \in \mathbb{R}\), initialization timestamp \(t_0\) and last updated timestamp \(t\). With a radius property, a surfel can represent a local flat surface around a given a position \(\bp\). 

Even though surfel fusion is fast with efficient implementation running on GPU, it takes a considerable amount of time. To speed up the efficiency, we leverage the advantage that surfels can easily be moved rigidly in space. We fuse \(k\) consecutive frames scene fragments as basic blocks and transform them based on optimized camera trajectories. In this way, the fusion of updated camera pose estimates is approximated with transforming scene fragments to updated location. Thus final results are calculated more efficiently.

\subsection*{Fragment loop to frame loop conversion}
\label{sec:orgecd136f}
Since there are fewer scene fragments than frames, there is a need to convert scene fragment matches to camera frame loops. We do this by connecting a reference frame in one scene fragment and connecting it to all the frames of the other scene fragments and repeat for the other direction.

\section*{EXPERIMENTS}
\label{sec:orgf51ae63}
Extensive experiments are performed to evaluate our proposed method on two datasets: augmented ICL-NUIM~\cite{Choi2015} and SUN3D dataset~\cite{Xiao2013}. SMD is short for surface mean distance. 

\begin{table*}[!bht]
\caption{Performance difference with only key loops vs. all correct loops agreed by ground truth. SMD is short for surface mean distance.}
\label{tab_key_loop_vs_all_loop}
\begin{center}
\begin{tabular}{|l|l|l|l|l|l|l|}
\hline
 & \multicolumn{2}{l|}{Traj. RMSE} & \multicolumn{2}{l|}{SMD} & \multicolumn{2}{l|}{DMP} \\
\cline{2-7}
 & key loops & all loops & key loops & all loops & key loops & all loops \\
\hline
livingroom 1 & \textbf{0.082} & 0.175 & \textbf{0.027} & 0.059 & \textbf{36.9} & 122.8 \\
livingroom 2 & \textbf{0.037} & 0.203 & \textbf{0.012} & 0.080 & \textbf{19.5} & 85.2 \\
office 1     & \textbf{0.051} & 0.096 & \textbf{0.020} & 0.046 & \textbf{143.5} & 200.9 \\
office 2     & \textbf{0.036} & 0.085 & \textbf{0.014} & 0.024 & \textbf{110.5} & 373.2 \\
\hline
Average      & \textbf{0.052} & 0.140 & \textbf{0.018} & 0.052 & \textbf{77.6} & 195.5 \\
\hline
\end{tabular}
\end{center}
\end{table*}

\subsection*{Augmented ICL-NUIM dataset}
\label{sec:org44ccd62}

We run experiments on the Augmented ICL-NUIM dataset~\cite{Choi2015}. This dataset is a synthetic dataset with ground-truth surface models and camera trajectories. The dataset has four data sequences of RGB-D data. For each sequence, there are merged scene fragments available with ground truth registration results. For this dataset, our baseline method is CZK~\cite{Choi2015} which is published in the same work as the Augmented ICL-NUIM dataset. 

\begin{table}[htbp]
\caption{Recall and precision performance on loops before and after loop filtering or sifting. \label{tab_icl_pr}}
\centering
\begin{tabular}{lccc}
\hline
Recall/Precision (\%) & Registration & ZGK & Ours\\
\hline
livingroom 1 & \textbf{61.2} / 27.2 & 57.6 / 95.1 & 5.5 / \textbf{100}\\
livingroom 2 & \textbf{49.7} / 17.0 & \textbf{49.7} / 97.4 & 3.9 / \textbf{100}\\
office 1 & \textbf{64.4} / 19.2 & 63.3 / 98.3 & 2.8 / \textbf{100}\\
office 2 & \textbf{61.5} / 14.9 & 60.7 / \textbf{100} & 0.7 / \textbf{100}\\
\hline
\end{tabular}
\end{table}

Experiments are conducted to evaluate the loop sifting and majorization performance of the proposed method. Performance is evaluated based on precision and recall of loops detected and remaining. Results are reported in table~\ref{tab_icl_pr}.
We can see that our method gets 100\% percent precision, which is desired. You may notice that the recall reduced dramatically after sifting. The decrease is not because of the strict requirement but because many loops are not very useful. We note that the remaining loops are the core ones that matter most for a better reconstruction quality, which we call it loop majorization. 

Many of the original loops are close to each other and connected accurately by ORB-SLAM tacking already. We prove this in another experiment that evaluates the trajectories and reconstructed 3D models of optimization with the key loops identified by our method and all the loops that agree with ground truth. Results are shown in table~\ref{tab_key_loop_vs_all_loop}. We can see that more loops do not improve performance instead decrease the performance. This is because some of the loops are not very precise. It will decrease accuracy if two loop regions are well connected originally.

\begin{figure}[bht]\centering
\subfloat[livingroom 1]{ \includegraphics[width=0.47\linewidth]{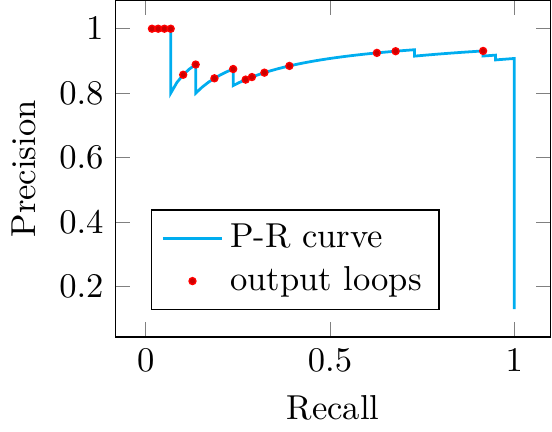}}
\subfloat[livingroom 2]{ \includegraphics[width=0.47\linewidth]{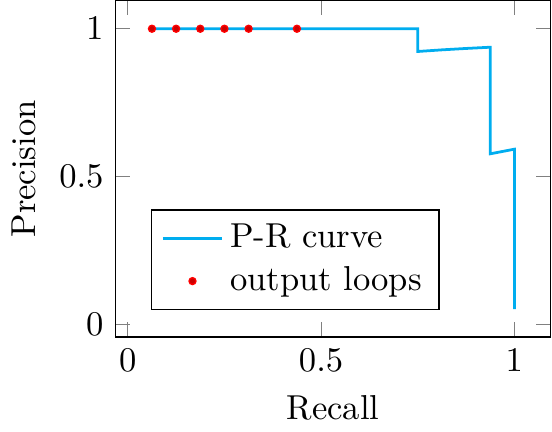}}\\
\subfloat[office 1]{ \includegraphics[width=0.47\linewidth]{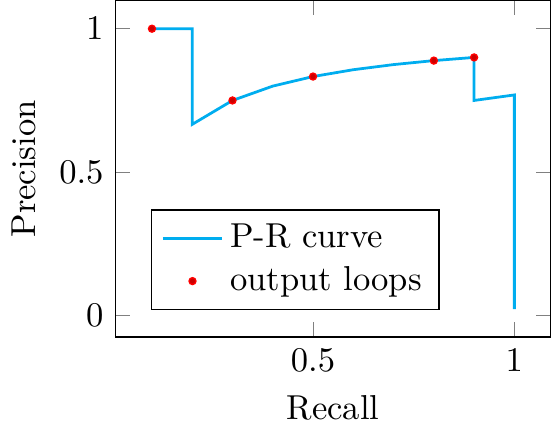}}
\subfloat[office 2]{ \includegraphics[width=0.47\linewidth]{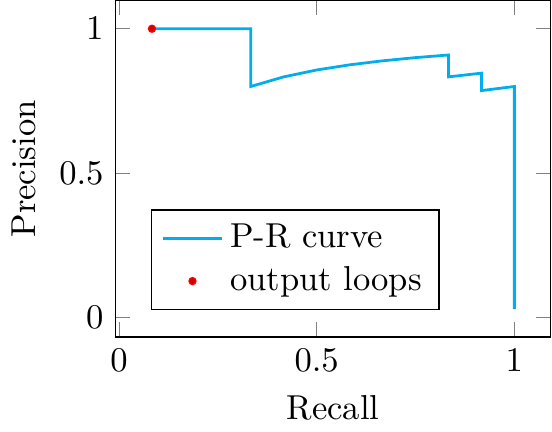}}\\
\caption{Precision and recall curves of ranking results in algorithm 1. The curves show a good ranking performance with exceptions. These false-positive loops that remained in the ranking are well handled by the last part, which tests and decides acceptance of each ranked loop. The positions of final output loops are also marked in red dots in the figures. We can notice that the algorithm performs smart selections on the ranking results because not all top ranking loops are selected.}\label{fig:nuim_PR}
\end{figure}

To further understand the proposed method, we draw precision-recall curves of the loop ranking results in the proposed algorithm in figure~\ref{fig:nuim_PR}. In the results, we can see the curves all starts from \(100\%\) precision. Then the curves keep on high precision values when recall increases. There are a few drop points, which means false positive loops. The majority of the false-positive loops are at the end of the list reflected by the sharp drops when recall reaches \(100\%\). These mean that the ranking has good performance with exceptions. These false-positive loops that remained in the ranking are well handled by the last part, which tests and decides acceptance of each ranked loop. It shows one more strength of our method: it decides the acceptance of correct loops without a single user parameter, even when the ratio of true/false positive loops are drastically different.

\begin{figure*}[!htbp]
  \centering
 \begin{tabular}{ccc}
  & Tracking only & Optimized with output loops\\
\rotatebox{90}{\qquad \quad maryland\_hotel3} & \subfloat {\centering\includegraphics[width=0.45\textwidth]{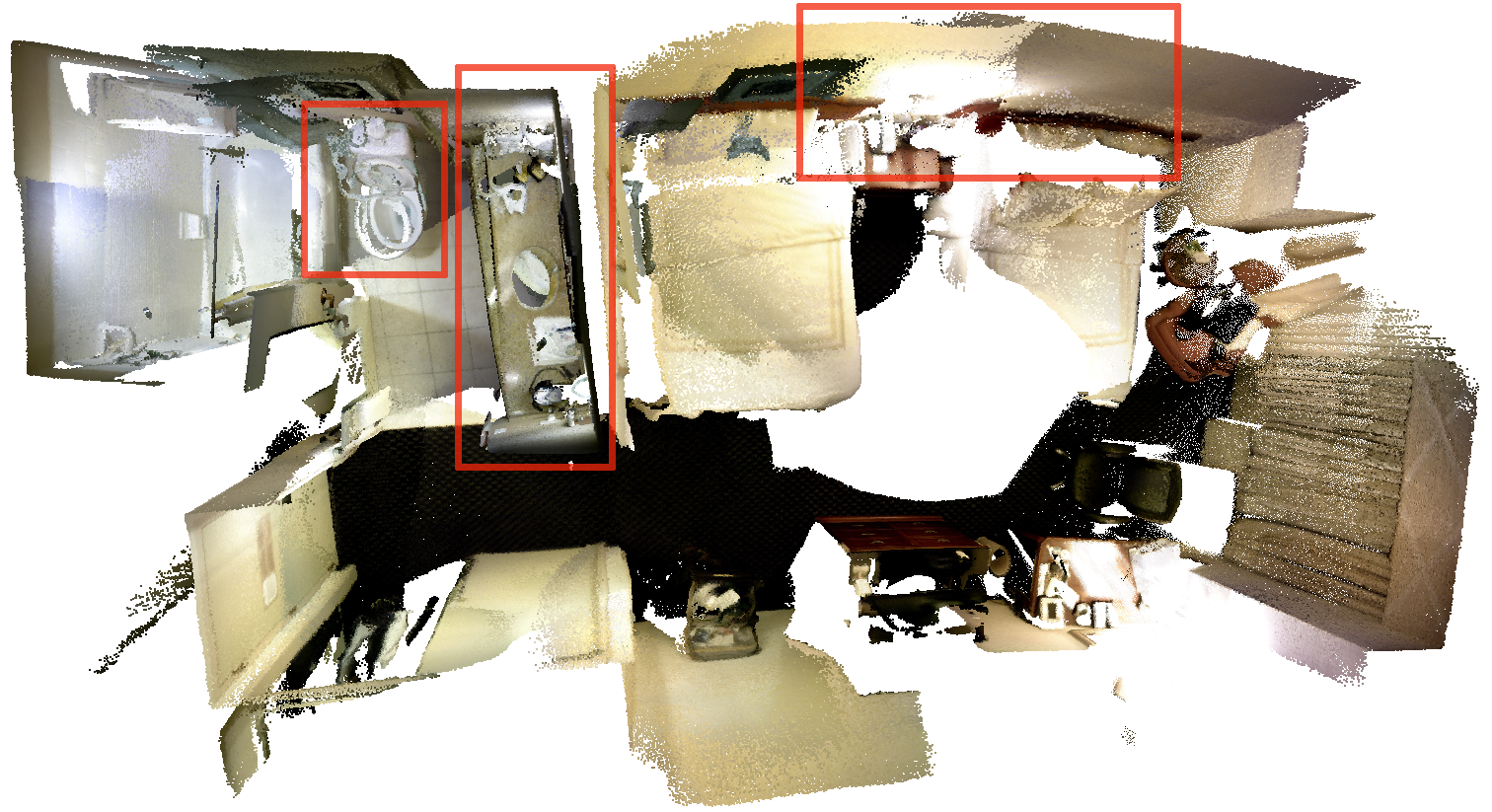}\label{fig:tabc}}&
   \subfloat{\centering\includegraphics[width=0.45\textwidth]{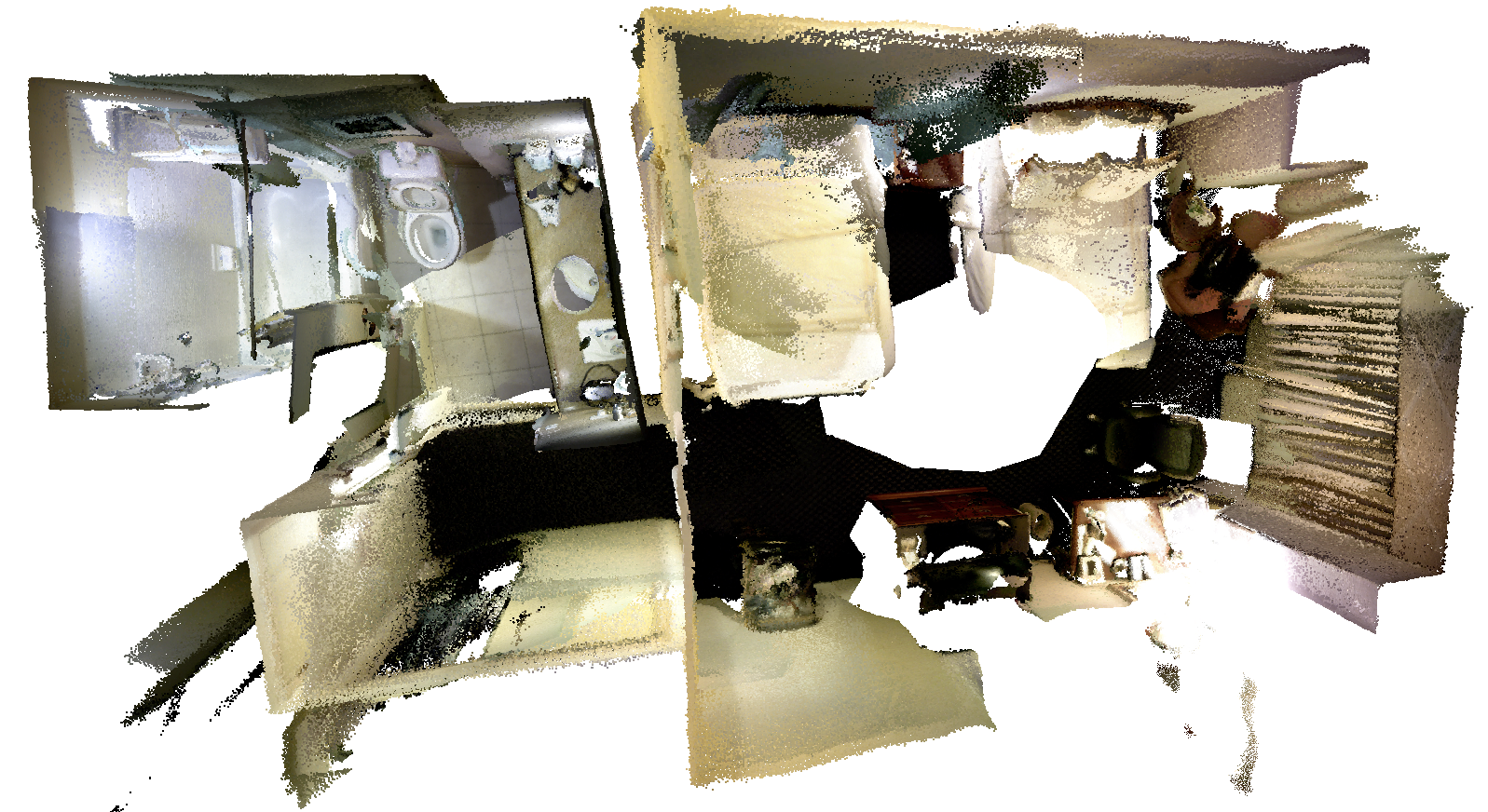}\label{fig:tabb}} \\
\rotatebox{90}{\qquad \qquad \qquad 76\_studyroom}& \subfloat{\centering\includegraphics[width=0.45\textwidth]{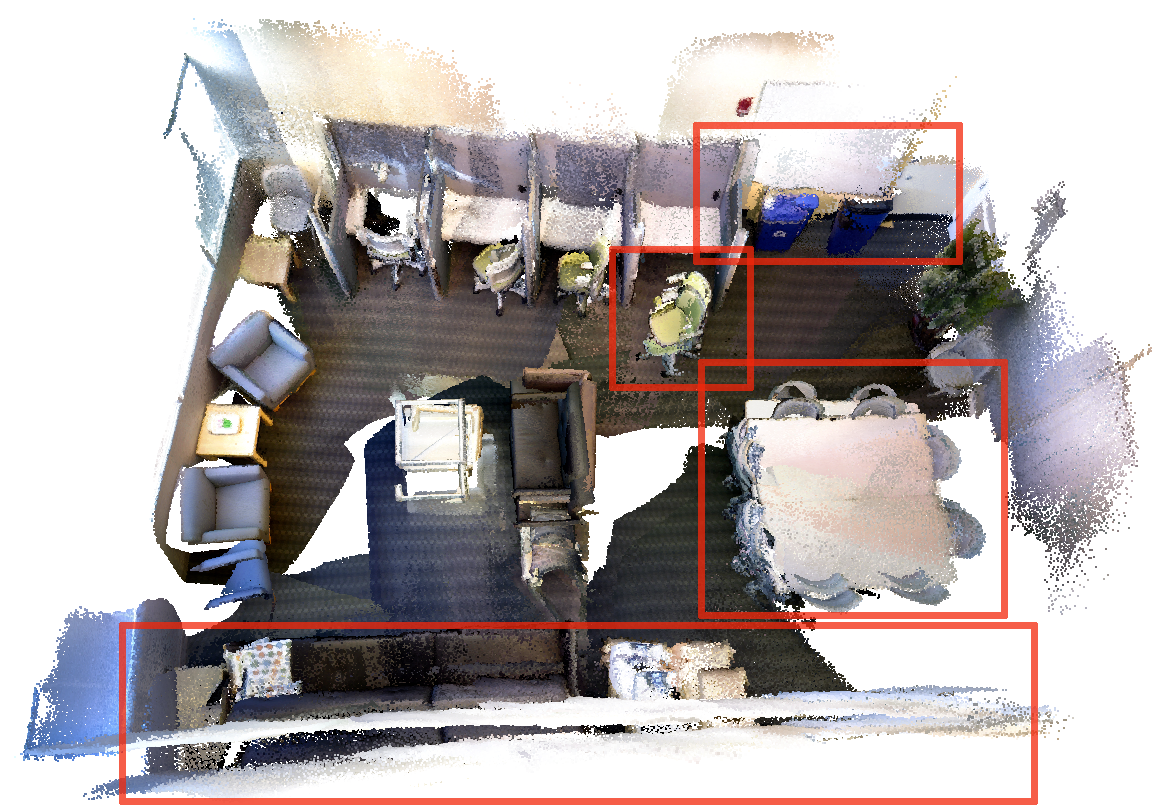}\label{fig:tabc}}&
 \subfloat{\centering\includegraphics[width=0.45\textwidth]{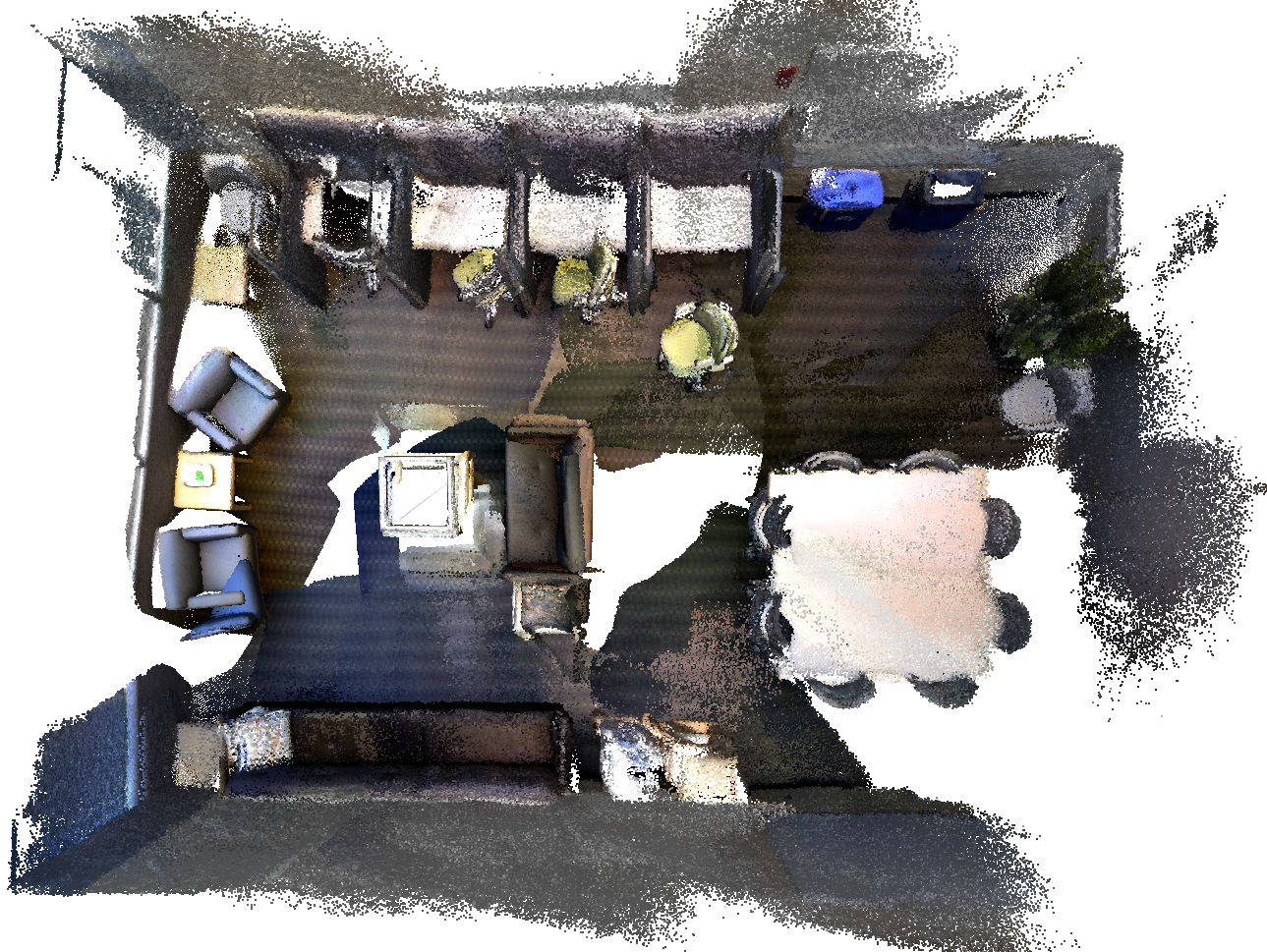}\label{fig:tabb}} \\
 \rotatebox{90}{\quad \quad \quad \quad mit\_32\_d507} & \subfloat{\centering\includegraphics[width=0.45\textwidth]{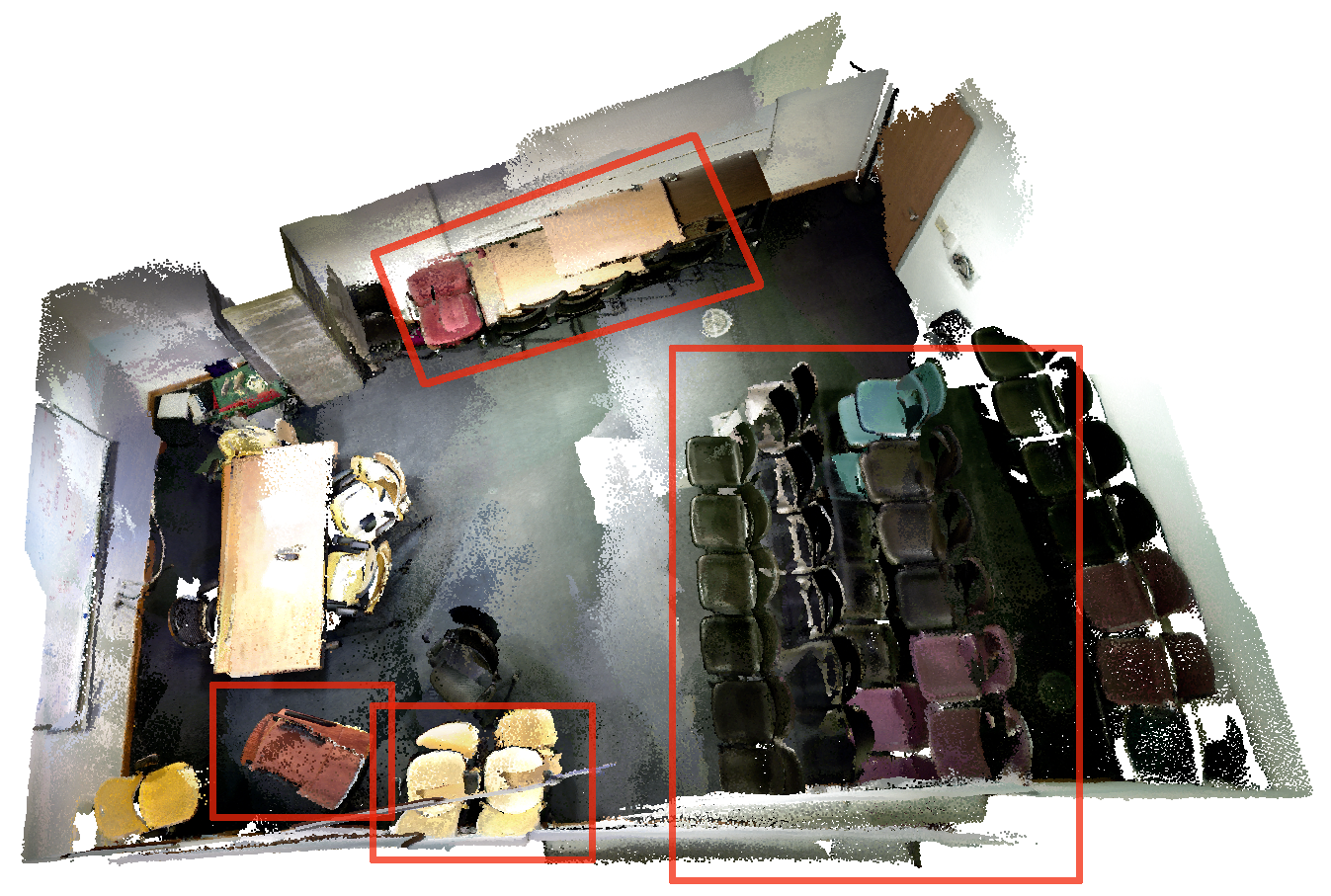}\label{fig:taba}}&
 \subfloat{\centering\includegraphics[width=0.45\textwidth]{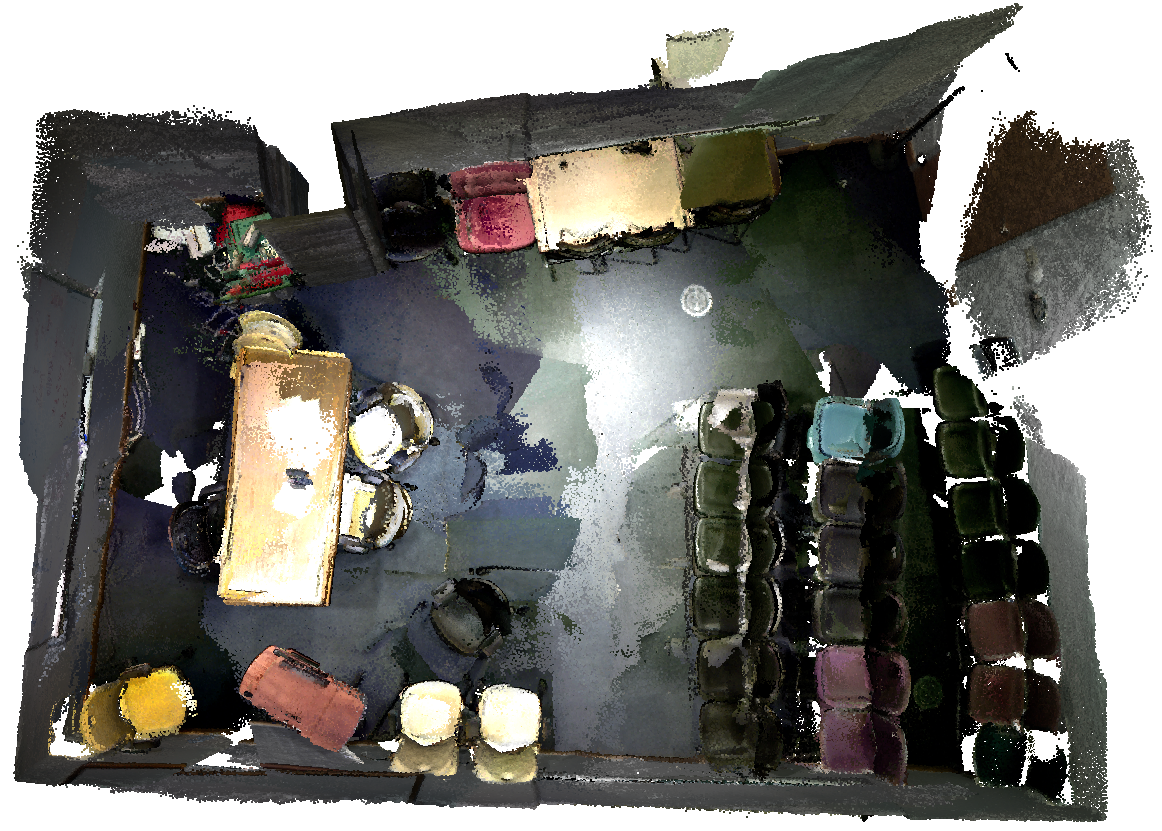}\label{fig:taba2}}
 \end{tabular}
 \caption{Results on sequences of the SUN3D dataset. The left column shows the results of the tracking part. The right column shows the results after our loop optimization. We highlight the mismatches in tracking results so that differences are easier compared.}
 \label{fig:sun3d_results}
 \end{figure*}

\subsection*{SUN3D dataset}
\label{sec:org4080e32}
The SUN3D dataset~\cite{Xiao2013} is a large-scale RGB-D database. It contains many data sequences captured at many places. Among them, there are eight sequences (listed here: \url{http://sun3d.cs.princeton.edu/listNow.html}) that are labeled with object annotations and widely used for evaluating SLAM and 3D reconstruction systems. We follow this common practice and run experiments on these sequences. For sequences \texttt{harvard\_c5}, \texttt{harvard\_c6} and \texttt{harvard\_c8}, there is no loops detected on top of tracking. So we do not include results for them.

\begin{table*}
\caption{DMP Performance difference of different methods on different sequences}
\label{tab_sun3d_dmp}
\begin{center}
\begin{tabular}{lcccccc}
\hline
 & SUN3D & ORB SLAM2 & CZK & BundleFusion & Tracking & Ours\\
\hline
mit\_32\_d507 & 573.95 & 750.60 & 334.17 & 441.15 & 904.25 & \textbf{296.59}\\
maryland\_hotel3 & 145.85 & 107.50 & 108.91 & 128.86 & 111.83 & \textbf{96.56}\\
76\_studyroom & 448.84 & 1191.40 & 282.22 & 256.07 & 358.93 & \textbf{193.94}\\
mit\_dorm\_next & 46.38 & 51.88 & 734.50 & 944.81 & 87.04 & \textbf{30.16}\\
mit\_lab\_hj & 180.49 & 162.98 & 244.86 & 207.94 & \textbf{155.86} & 199.57\\
\hline
\end{tabular}
\end{center}
\end{table*}

\begin{table}[!htb]
\caption{Number of loops before and after loop sifting and majorization \label{fig_sun3d_loop_num}}
\centering
\begin{tabular}{lcc}
\hline
Number of loops & before & after\\
\hline
mit\_32\_d507 & 2135 & 34\\
maryland\_hotel3 & 224 & 6\\
76\_studyroom & 442 & 7\\
mit\_dorm\_next & 621 & 6\\
mit\_lab\_hj & 219 & 7\\
\hline
\end{tabular}
\end{table}

Quantitatively, we evaluate the DMP metric of different methods and report them in table~\ref{tab_sun3d_dmp}.  We compared with four different methods: 1) CZK, which is an offline method that targets the best surface reconstruction quality; 2) SUN3D, which is an offline method that adds manual object labeling as a source of loop closure; 3) ORB-SLAM2, which is a well known good SLAM that also has a tracking part in our system; 4) BundleFusion which is a well-engineered real-time dense SLAM system. 
The DMP metric evaluates that the proposed method makes reliable improvements on its initial start point: tracking result. Most importantly, it outperforms most methods. 
To understand the proposed method, we also include the number of loop detections and the number of loops that pass our algorithm, shown in table~\ref{fig_sun3d_loop_num}.

Qualitative, we show results in the form of screenshots of reconstructed 3D models in figure~\ref{fig:sun3d_results}. 
In figure~\ref{fig:sun3d_results}, we highlight the mismatches in tracking results so that reader can better compare them with our results.  SUN3D data sequences are scanned with very loopy motion in some areas but only once for some other scene parts, thus is considered difficult to process. Readers may refer to \url{http://redwood-data.org/indoor/models.html} for CZK results and \url{https://graphics.stanford.edu/projects/bundlefusion/recons.html} for BundleFusion results for visual comparison.

\section*{CONCLUSION}
\label{sec:orga24889d}

In this work, an algorithm that can sift and majorize loop detections is proposed. The algorithm tests and decides the acceptance of each loop without a single user-defined threshold.  Experiments are conducted on public datasets, including the Augmented ICL-NUIM dataset and the SUN3D dataset. Results show that the proposed method outperforms the state-of-the-art method. It can find key loops with \(100\%\) precision and eliminate significant mismatches when processing SUN3D sequences.

\bibliographystyle{asmems4}
\bibliography{../../../Dropbox/library/gxzbibtex}

\begin{thebibliography}{10}

\bibitem{MurArtal2017}
Mur-Artal, R., and Tardos, J.~D., 2017.
\newblock ``{ORB}-{SLAM}2: {A}n {O}pen-{S}ource {SLAM} {S}ystem for
  {M}onocular, {S}tereo, and {RGB}-{D} {C}ameras''.
\newblock {\em IEEE Transactions on Robotics, \textbf{ 33}}(5), Oct.,
  pp.~1255--1262.

\bibitem{Endres2014}
Endres, F., Hess, J., Sturm, J., Cremers, D., and Burgard, W., 2014.
\newblock ``3-{D} {M}apping {W}ith an {RGB}-{D} {C}amera''.
\newblock {\em {IEEE} Transactions on Robotics, \textbf{ 30}}(1), Feb.,
  pp.~177--187.

\bibitem{Klein2007}
Klein, G., and Murray, D., 2007.
\newblock ``Parallel {T}racking and {M}apping for {S}mall {AR} {W}orkspaces''.
\newblock In Proc. of the 2007 6th {IEEE} and {ACM} International Symposium on
  Mixed and Augmented Reality, {IEEE}.

\bibitem{Dai2017e}
Dai, A., Nie{\ss}ner, M., Zollh{\"{o}}fer, M., Izadi, S., and Theobalt, C.,
  2017.
\newblock ``{B}undle{F}usion: {R}eal-{T}ime {G}lobally {C}onsistent 3{D}
  {R}econstruction {U}sing {O}n-the-{F}ly {S}urface {R}eintegration''.
\newblock {\em ACM Transactions on Graphics, \textbf{ 36}}(3), May, pp.~1--18.

\bibitem{Liang2019}
Liang, H.-J., Sanket, N.~J., Fermuller, C., and Aloimonos, Y., 2019.
\newblock ``{SalientDSO: Bringing Attention to Direct Sparse Odometry}''.
\newblock {\em IEEE Transactions on Automation Science and Engineering,
  \textbf{ 16}}(4), Oct., pp.~1619--1626.

\bibitem{Zubizarreta2020}
Zubizarreta, J., Aguinaga, I., and Montiel, J. M.~M., 2020.
\newblock ``{Direct Sparse Mapping}''.
\newblock {\em IEEE Transactions on Robotics, \textbf{ 36}}(4), Aug.,
  pp.~1363--1370.

\bibitem{Engel2014}
Engel, J., Sch{\"{o}}ps, T., and Cremers, D., 2014.
\newblock ``{LSD}-{SLAM}: Large-scale direct monocular {SLAM}''.
\newblock In {\em Proc. of the Computer Vision {\textendash} {ECCV} 2014}.
  Springer International Publishing, pp.~834--849.

\bibitem{Newcombe2011}
Newcombe, R.~A., Izadi, S., Hilliges, O., Molyneaux, D., Kim, D., Davison,
  A.~J., Kohi, P., Shotton, J., Hodges, S., and Fitzgibbon, A., 2011.
\newblock ``{KinectFusion}: Real-time dense surface mapping and tracking''.
\newblock In Proc. of the 2011 10th IEEE International Symposium on Mixed and
  Augmented Reality, pp.~127--136.

\bibitem{Whelan2015}
Whelan, T., Leutenegger, S., Moreno, R.~S., Glocker, B., and Davison, A., 2015.
\newblock ``{{ElasticFusion}: Dense {SLAM} Without A Pose Graph}''.
\newblock In Proc. of the Robotics: Science and Systems {XI}, Robotics: Science
  and Systems Foundation.

\bibitem{Concha2016}
Concha, A., Loianno, G., Kumar, V., and Civera, J., 2016.
\newblock ``Visual-inertial direct {SLAM}''.
\newblock In Proc. of the 2016 {IEEE} International Conference on Robotics and
  Automation ({ICRA}), {IEEE}.

\bibitem{Usenko2016}
Usenko, V., Engel, J., Stuckler, J., and Cremers, D., 2016.
\newblock ``Direct visual-inertial odometry with stereo cameras''.
\newblock In Proc. of the 2016 {IEEE} International Conference on Robotics and
  Automation ({ICRA}), {IEEE}.

\bibitem{Niesner2014}
Nie{\ss}ner, M., Dai, A., and Fisher, M., 2014.
\newblock ``Combining {I}nertial {N}avigation and {ICP} for {R}eal-time {3D}
  {S}urface {R}econstruction.''.
\newblock In Proc. of the Eurographics (Short Papers), Citeseer, pp.~13--16.

\bibitem{ZihaoZhu2017}
Zihao~Zhu, A., Atanasov, N., and Daniilidis, K., 2017.
\newblock ``{Event-Based Visual Inertial Odometry}''.
\newblock In Proc. of The IEEE Conference on Computer Vision and Pattern
  Recognition (CVPR).

\bibitem{Dong2017}
Dong, J., Fei, X., and Soatto, S., 2017.
\newblock ``Visual-inertial-semantic scene representation for {3D} object
  detection''.
\newblock In Proc. of The IEEE Conference on Computer Vision and Pattern
  Recognition (CVPR).

\bibitem{Liu2018b}
Liu, H., Chen, M., Zhang, G., Bao, H., and Bao, Y., 2018.
\newblock ``{ICE-BA}: Incremental, consistent and efficient bundle adjustment
  for visual-inertial {SLAM}''.
\newblock In Proc. of The IEEE Conference on Computer Vision and Pattern
  Recognition (CVPR).

\bibitem{Cadena2016}
Cadena, C., Carlone, L., Carrillo, H., Latif, Y., Scaramuzza, D., Neira, J.,
  Reid, I., and Leonard, J.~J., 2016.
\newblock ``Past, {P}resent, and {F}uture of {S}imultaneous {L}ocalization and
  {M}apping: {T}oward the {R}obust-{P}erception {A}ge''.
\newblock {\em IEEE Transactions on Robotics, \textbf{ 32}}(6), Dec.,
  pp.~1309--1332.

\bibitem{GalvezLopez2012}
Galvez-L{\'{o}}pez, D., and Tardos, J.~D., 2012.
\newblock ``Bags of binary words for fast place recognition in image
  sequences''.
\newblock {\em IEEE Transactions on Robotics, \textbf{ 28}}(5), Oct.,
  pp.~1188--1197.

\bibitem{Cummins2008}
Cummins, M., and Newman, P., 2008.
\newblock ``{FAB-MAP}: {P}robabilistic {L}ocalization and {M}apping in the
  {S}pace of {A}ppearance''.
\newblock {\em International Journal of Robotics Research, \textbf{ 27}}(6),
  pp.~647--665.

\bibitem{Kummerle2011}
Kummerle, R., Grisetti, G., Strasdat, H., Konolige, K., and Burgard, W., 2011.
\newblock ``G2o: A general framework for graph optimization''.
\newblock In Proc. of the 2011 {IEEE} International Conference on Robotics and
  Automation, {IEEE}.

\bibitem{Zhang2021a}
Zhang, G., and Chen, Y.
\newblock A metric for evaluating 3{D} reconstruction and mapping performance
  with no ground truthing.
\newblock arXiv:2101.10402.

\bibitem{Lee2013}
Lee, G.~H., Fraundorfer, F., and Pollefeys, M., 2013.
\newblock ``Robust pose-graph loop-closures with expectation-maximization''.
\newblock In Proc. of the 2013 {IEEE}/{RSJ} International Conference on
  Intelligent Robots and Systems, {IEEE}.

\bibitem{Sunderhauf2013}
Sunderhauf, N., and Protzel, P., 2013.
\newblock ``{Switchable constraints vs. max-mixture models vs. {RRR} - A
  comparison of three approaches to robust pose graph {SLAM}}''.
\newblock In Proc. of the 2013 {IEEE} International Conference on Robotics and
  Automation, {IEEE}.

\bibitem{Agarwal2013}
Agarwal, P., Tipaldi, G.~D., Spinello, L., Stachniss, C., and Burgard, W.,
  2013.
\newblock ``Robust map optimization using dynamic covariance scaling''.
\newblock In Proc. of the 2013 {IEEE} International Conference on Robotics and
  Automation, {IEEE}.

\bibitem{Sunderhauf2012}
Sunderhauf, N., and Protzel, P., 2012.
\newblock ``Switchable constraints for robust pose graph {SLAM}''.
\newblock In Proc. of 2012 {IEEE}/{RSJ} International Conference on Intelligent
  Robots and Systems, {IEEE}.

\bibitem{Choi2015}
Choi, S., Zhou, Q.-Y., and Koltun, V., 2015.
\newblock ``Robust reconstruction of indoor scenes''.
\newblock In Proc. of the 2015 {IEEE} Conference on Computer Vision and Pattern
  Recognition ({CVPR}), {IEEE}.

\bibitem{Xiao2013}
Xiao, J., Owens, A., and Torralba, A., 2013.
\newblock ``{SUN}3{D}: {A} {D}atabase of {B}ig {S}paces {R}econstructed {U}sing
  {SfM} and {O}bject {L}abels''.
\newblock In Proc. of the 2013 {IEEE} International Conference on Computer
  Vision, {IEEE}.

\bibitem{Pfister2000}
Pfister, H., Zwicker, M., van Baar, J., and Gross, M., 2000.
\newblock ``Surfels: {S}urface {E}lements as {R}endering {P}rimitives''.
\newblock In Proc. of the 27th annual conference on Computer graphics and
  interactive techniques - {SIGGRAPH} {\textquotesingle}00, {ACM} Press.

\end{thebibliography}
\end{document}